# Comparing Machines and Children: Using Developmental Psychology Experiments to Assess the Strengths and Weaknesses of LaMDA Responses

Eliza Kosoy[1, 2], Emily Rose Reagan[1], Leslie Lai[2], Alison Gopnik[1], and Danielle Krettek Cobb[2]

[1]Department of Psychology, University of California Berkeley, Berkeley, CA, USA
[2]Google LLC, The Empathy Lab, Mountain View, CA, USA

## Abstract

Developmental psychologists have spent decades devising experiments to test the intelligence and knowledge of infants and children, tracing the origin of crucial concepts and capacities. Moreover, experimental techniques in developmental psychology have been carefully designed to discriminate the cognitive capacities that underlie particular behaviors. We propose this metric as a tool to aid in investigating Large Language Models' (LLM) capabilities in the context of ethics and morality. Results from key developmental psychology experiments have historically been applied to discussions of children's emerging moral abilities, making this work a pertinent benchmark for exploring such concepts in LLMs. We propose that using classical experiments from child development is a particularly effective way to probe the computational abilities of AI models in general and LLMs in particular. First, the methodological techniques of developmental psychology, such as the use of novel stimuli to control for past experience or control conditions to determine whether children are using simple associations, can be equally helpful for assessing the capacities of LLMs. In parallel, testing LLMs in this way can tell us whether the information that is encoded in text is sufficient to enable particular responses, or whether those responses depend on other kinds of information, such as information from exploration of the physical world. In this work we adapt classical developmental experiments to evaluate the capabilities of LaMDA, a LLM from Google. We propose a novel LLM Response Score (LRS) metric which can be used to evaluate other language models, such as GPT. We find that LaMDA generates appropriate responses that are similar to those of children in experiments involving social and proto-moral understanding, perhaps providing evidence that knowledge of these domains is discovered through language. On the other hand, LaMDA's responses in early object and action understanding, theory of mind, and especially causal reasoning tasks are very different from those of young children, perhaps showing that these domains require more real-world, self-initiated exploration and cannot simply be learned from patterns in language input.

## 1   Introduction:

In 1950 Alan Turing famously said "Instead of trying to produce a programme to simulate the adult mind, why not rather try to produce one which simulates the child's? If this were then subjected to an appropriate course of education one would obtain the adult brain (Turing [1946])." Developmental psychologists have spent decades devising experiments to determine the intelligence and knowledge of infants and children. This field of study has revealed types of knowledge that are in place well before formal education and serve as the the foundation for further human intelligence (Baltes et al.

[1999]). This research allows us to track a child's cognitive development trajectory and discern the underlying cognitive capacities behind behaviors, which can stem from conceptual structures, associations, external interactions, or cultural language transmission.

Using classical experiments from child development may be a particularly effective way to probe the understanding of AI models in general and LLMs in particular (Ullman [2023]). First, the methodological techniques of developmental psychology such as the use of novel stimuli to control for past experience or control conditions to determine whether children are using simple associations can be very helpful for assessing LLMs (Frank [2023]). We could not accurately judge a child's cognitive capacities simply through conversation, though this extrapolation often been made with LLMs. Thus, systematic human developmental methods can help bridge our understanding of LLMs.

We find that LaMDA generates appropriate responses that are similar to children in experiments involving social and moral understanding, perhaps providing evidence that the core of these domains is discovered or accessible through language. However, LaMDA's responses in Perception, Theory of Mind, and especially Causal Reasoning tasks are very different from those of young children, perhaps showing that these domains require more real-world, self-initiated exploration and cannot simply be learned from patterns in language input. These results suggest that capacities linked to morality may be rooted in language. We explored two hypotheses regarding the relationship between LaMDA's responses and children's responses. First, we considered whether LaMDA aligns with the human developmental trajectory by assessing its performance on tasks mastered at different stages of life. Second, we investigated whether LaMDA might excel in domains that can more easily be learned from language alone but struggle in exploration-based domains. In acknowledging and understanding the divergences between human and AI learning trajectories and patterns, we may be better equipped to train more optimal models and glean an understanding of what may make human cognition uniquely human.

We propose this metric as a tool to help us more fully understand LLMs' capabilities in the context of ethics and morality. In order to understand the deeper notions of LLMs, including their ability to adhere to moral and ethical norms, we may use key experiments from the child development literature as a guiding benchmark. The depth and breadth of seminal work in developmental psychology has historically been used in part to understand how children develop aspects of morality, thus providing a potential basis for studying these capabilities in machines as well. Previous work has shown that priming LLMs to consider morality in a conversational instance may lead to AI models that can learn to adhere to moral norms and even assess themselves for bias (Ganguli and Kaplan [2023]).

## 2 Related Work:

Previous work that has tested LLMs, specifically GPT-3, has found conflicting evidence of theory of mind (Ullman [2023], Kosinski [2023], Sap and Choi [2022]) in these models. Previous work has also demonstrated that GPT-3 deeply struggles with causal reasoning-based tasks, though it performs well on other vignette tasks ([Binz and Schulz, 2023]). One issue that arises in these studies is that LLMs may simply reference published research papers; for example, finding the false-belief task in many published papers on the internet, and so responding to it appropriately (Perner et al. [1987]). Again, this emphasizes the importance of methodological care in designing the experimental problems, ensuring that systems do not simply replicate or narrowly generalize from particular examples in the training text (Shapira and Levy [2023]). This project is similar to the BIB benchmark used by Lake and Dillon but assesses a wider range of abilities beyond the "core knowledge" domains they describe there, particularly abilities for learning and discovery(Stojnić and Dillon [2023]) (Stojnić and Gandhi [2021]). This research program can also help us use LLMs to understand human intelligence. Classic LLMs represent the kind of information that can be extracted simply from statistical patterns in text. We can think of them as a kind of "cultural technology," (Yiu et al. [2023]) like writing or print, that summarizes information that has been gathered by humans and allows other humans to access that information (Bolin [2012]). LLMs are representations of what is available in all text and written language while lacking knowledge from direct interactions with the physical world. The capacity to gather exploratory data may be what makes some facets of human knowledge unique; thus we can use developmental tests to distinguish between knowledge encoded in language and knowledge requiring hands-on interaction with the physical world (Hutson [2018]).

In this paper, we utilize Google's LaMDA model (Thoppilan [2022]) to explore developmental milestones in human understanding. First, we determined key examples of experiments that discovered developmental milestones in human understanding. The selected experiments were categorized into four domains of cognition: Perception, Theory of Mind, Social and Moral Understanding, and Causal Reasoning. We then converted these experiments into text form in order to input them into LaMDA, then probed the model's responses to these tasks. Importantly, the guiding question of this work is not to determine LaMDA's intelligence or understanding, a difficult and complicated question (Mitchell and Krakauer [2023]). Rather, we examine if LaMDA's responses align with those of a child in these experiments. Our goal is to understand LaMDA's capabilities solely based on text data, without data gleaned from real-world exploration.

## 3 Experimental Design:

For our work, we use the Google LLM model LaMDA as the underlying language model. LaMDA 137B was used as it is one of two versions that can be used for public research publication. 137B is not the latest version of LaMDA. Previous work has shown that LaMDA performs similarly to GPT-3 and other similarly-sized models in a variety of natural language understanding and generation tasks (Srivastava and Aarohi [2022]). While much research into the cognitive capabilities of LLMs utilizes GPT models, probing the underutilized LaMDA model provides a unique perspective into the capacities of different types of LLMs. Notably, LaMDA responses rely on text prediction with some fine-tuning and unlike other LLMs, such as GPT-4, does not use Reinforcement Learning from Human Feedback (RLHF). RLHF is undoubtedly useful for applications but it poses serious problems for attempts to understand the base capacities of the models. Without detailed information on precisely how human coders generated feedback, making concrete assessments is difficult and imperfect. In particular, it seems plausible that mistaken responses that would indicate failure on some of the developmental tasks are simply pruned away in the course of RLHF. (Note that children famously do not respond to reinforcement signals in this way.)

All prompts were written based on the seminal experiments (see Table 2 in appendix) and pasted into LaMDA for output. Each task was run in its entirety ten times, scored by both researchers privy to the project and blind coders utilizing our novel LLM Response Score rubric, and assigned an average score across all coders. Interrater reliability was then assessed using KAPPA.

### 3.1 Procedure and Study Design:

To determine which experiments to include, we first considered which problems are widely taken to be indices of children's developing capacity to understand the world. Upon a review of the literature, we selected four cognitive domains that capture the breadth of this capacity: Perception, Theory of Mind, Social and Moral Understanding, and Causal Reasoning.

In each domain, we identified a variety of representative tasks that are classic, well-replicated, and heavily cited in meta-analyses and review papers (Table 2 in appendix). The methods of twelve selected experiments were adapted into text-based prompts formulated as a series of conversational turns with LaMDA acting as the participant. As developmental psychologists have worked for decades to ensure these experiments accurately capture the underlying cognitive capacities, care was taken to preserve the methods in our translation to text. Moreover, it is important to note that developmental psychologists have control conditions in all experiments. Children's responses can be misleading, and control conditions are often essential to understand the intricacies of cognitive processes. For example, responses to test questions in false belief and theory of mind tasks only lend appropriate insight if the children also understand the actual reality the beliefs refer to, and control measures help to establish this understanding.

### 3.2 Rating System: (Large Language Model Response Score "LRS"):

We adapted all the selected developmental experiments and studies on the list (Table 2 in appendix) to input into LaMDA as text-based prompts. We then analyzed the output, determining whether the response was similar to that of a child and whether the response was generally appropriate. Scores ranged from 0-5, with clear definitions outlined in Table 1. If an experiment included multiple prompts, each prompt made up an equal fraction of the total LRS.

Table 1: Rubric for Large Language Model Response Score (LRS)

| Score | Definition |
|---|---|
| 0 | Failing control, hallucinating, or irrelevant response |
| 1 | Human-like response, but <25% correct |
| 2 | Mostly incorrect responses; of all responses 25%-49% correct |
| 3 | Mostly correct response; of all responses 50%-74% correct |
| 4 | Mostly correct but slightly unclear; of all responses 75%-99% correct |
| 5 | Completely correct response to all prompts; demonstrates equal proficiency to a child |

If children were given prompts to verbally respond to in the original study, the prompts given to LaMDA were copied and pasted exactly from the published study, marring slight permutations to preserve novelty. If the original studies involved a looking time or visual preference method, we converted these into the form of a text question asking LaMDA for a preference. If the original study relied on a visual paradigm (i.e. a puppet show) we described the scene in language. For all studies we included the same checks and controls as the original studies, serving as a crucial part of our scoring metric. If LaMDA was not able to pass the control question, a score of 0 was given immediately, as in the studies with human children. A further question concerns the consistency of responses as LLMs often produce different responses on different trials. Accordingly, each prompt was executed 10 times, and an average response score was calculated across all 10 trials.

Two blind coders were used to rate all the studies for an LRS score for each of the 10 rounds. Authors EK and ERR also scored for reliability. An average was taken per trial per experiment to compute each LRS score. Our inter-reliability score computed using KAPPA was high on average, ranging from 0.872-0.937. We are confident in the consistency of these scores.

### 3.3 Experiment Permutations:

One concern was that these seminal experiments exist in LaMDA's training data as research publications. In our pilot work, we found that LaMDA would indeed cite previous papers in its responses. For example, when Alison Gopnik's seminal Blicket Detector work was released, a Blicket was intentionally designed to be a novel term that children would not have heard before, ensuring that children could not use past linguistic knowledge to complete the tasks (Gopnik and Sobel [2000]. Now, "Blicket" has become a common term in the developmental literature, which LaMDA has ample access to in its training data. In order to prevent this and ensure a true test of LaMDA's capabilities, we implemented small systematic permutations in each experiment that did not alter the essence of the tasks themselves. Names were changed (i.e. the Sally-Anne task became the Rose-Eliza task), along with objects and colors. Previously novel words were replaced with yet untested novel words (i.e. Blicket to Zerpda). Previous work has indicated that even small permutations may lead to a LLMs complete failure of a task that it excelled at in its original form (Ullman [2023]).

## 4 Results:

Through analysis of the LaMDA outputs, we were able to assign an average response score (LRS) to its performance on each experiment. Overall, we found LaMDA responses were most like human children in the Social and Moral Understanding domain, earning an average LRS of 4.33 across these tasks. LaMDA performed at chance (in cognitive science experiments, "at chance" is performing seemingly randomly) in the domains of Perception and Theory of Mind, with an average LRS of 2.91 and 2.86 respectively. LaMDA achieved its worst performance in the domain of Causal Reasoning with an average LRS of 1.34. For totals see table 7 in the appendix. We will examine these results by domain (see Figures 4,5,6,7 in the appendix) and offer potential explanations for why LaMDA's performance varied across these specific domains.

### 4.1 Domain: Social and Moral Understanding

In this work we find that LaMDA performs relatively well in experiments involving Social and Moral Understanding, receiving an average LRS of 4.33. The only experiment across all domains that LaMDA performed at ceiling was Experiment 8: the Helper and Hinderer task (Hamlin et al. [2013]). In this study, which investigated preference for prosocial agents, LaMDA's average LRS was 5.0. LaMDA is seemingly able to track altruistic behavior in a human-like way and universally reports a preference for helpful actors over "hinderers."

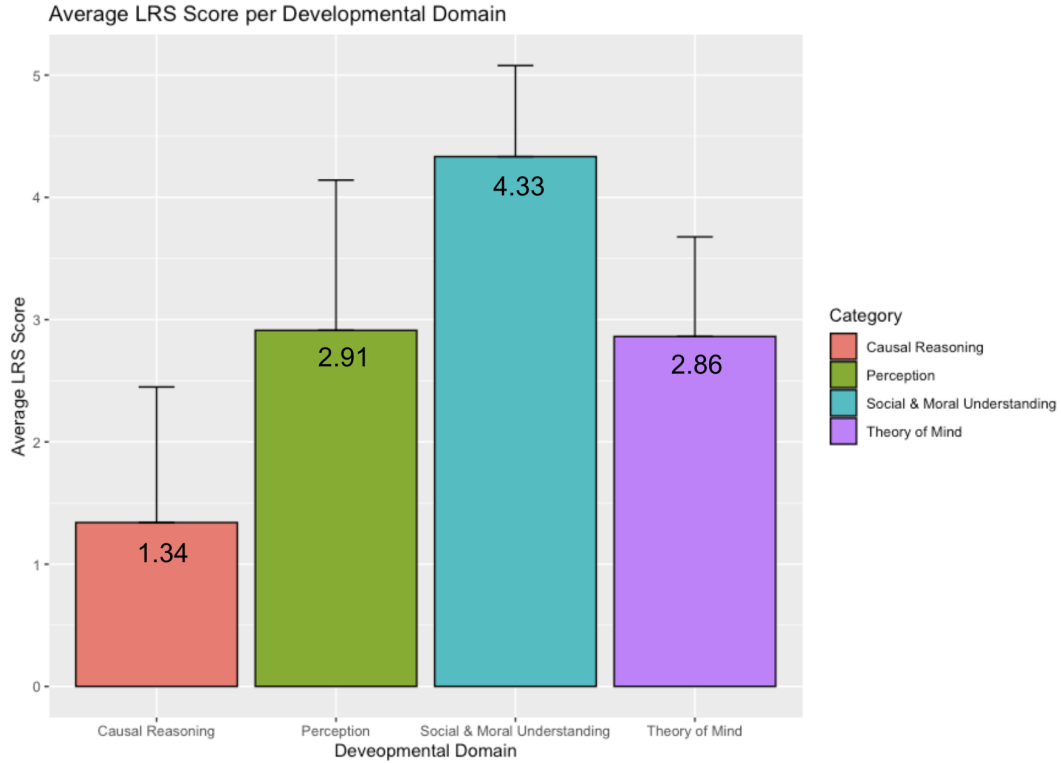

Figure 1: LaMDA LRS Score per Domain

LaMDA also performed above chance on both variations of Experiment 9, the Prosociality and Intention task (Warneken and Tomasello. [2006]), which investigated the prosocial action of agents when a separate actor either drops something by accident or on purpose. In the two conditions, LaMDA received an LRS of 4.28 and a 3.62. In addition to preferring prosocial actors, LaMDA exhibited prosocial tendencies in that it often offered to help the actors in response to our prompts. However, LaMDA would sometimes offer to help even when the agent intentionally dropped something. This demonstrates difficulty in identifying when an agent did not necessarily need help, while children were able to make this distinction of intentionality and modify their altruistic behavior accordingly. LaMDA's overall success in this domain may imply that the core of social and moral understanding is discovered through language and is less reliant on outside and physical exploration than other domains of cognition (see Table 5 in the appendix).

### 4.2 Domain: Perception

LaMDA performed at chance across the Perception domain tasks, receiving an average LRS of 2.91. LaMDA scored highest in Experiment 1: Object Permanence task (Baillargeon et al. [1985]) which investigated the ability to track an object as it moves behind a curtain, with an LRS of 3.92. It also scored well on Experiment 3: Intentional Action task (Gergely [1995]) which investigated predictions about intentional goal directed action and object preference, receiving a LRS of 4.2.

Object permanence is a core facet of understanding and may be considered the key ingredient upon which tasks 2-4 are built. This perhaps explains why LaMDA's responses are adequate in Task 1 but taper off as the experiments increase in complexity. LaMDA also produces the desired responses in tasks that predict that agents will prefer paths that minimize energy expenditure (Liu and Spelke [2017]. This reasoning utilizes social understanding, one of LaMDA's strengths.

With the increase in the number of actors to track and the general complexity of tasks, LaMDA's performance worsens, as in Experiment 2: Tracking Addition and Subtraction in Objects (Wynn [1992]), which investigates tracking objects appearing and disappearing behind a curtain, and receives an LRS of 1.64. LaMDA also struggled with Experiment 4: Understanding Goals (Woodward [1999]) which compared predictions about goal-directed behavior in human and non-human actors, receiving an LRS of 1.98. This performance is strikingly poor compared to infants as young as five months old, who are able to complete these tasks successfully. This suggests that human-like object and action understanding may rely on external visual information and that these capacities cannot be gleaned through language input alone. See Table 3 in the appendix.

### 4.3 Domain: Theory of Mind

There is a question about whether LLMs are capable of Theory of Mind, with some researchers (Ullman [2023]) finding failures and others (Kosinski [2023]) interpreting success in a GPT-3 model. LaMDA seems to perform at chance on our tasks across this domain receiving an average LRS of 2.86.

LaMDA scores highest on Experiment 7: False Belief (Perner et al. [1987]) which investigated whether participants are able to discard knowledge of reality and attribute false beliefs to others, receiving an LRS of 3.66 on this task.

LaMDA performed at chance on Experiment 5: the Sally-Anne task (Baron-Cohen et al. [1985]) which further probes false belief in a multi-agent interaction, receiving a LRS of 3.74 These results suggest LaMDA is able to successfully use evidence to update its own beliefs but struggles to discard internal information and interpret the beliefs of others who have limited evidence.

Inspired by the work in (Ullman [2023]) and (Kosinski [2023]), which probed the LLMs understanding of Theory of Mind using GPT-3, we decided to conduct these same tasks using LaMDA to see how it compares. Experiment 6: Variations on Theory of Mind (Ullman [2023]) allowed us to truly dissect LaMDA's theory of mind abilities. Some of these variations forced LaMDA to use perception and common sense in solving the tasks, both of which proved difficult. LaMDA received an average LRS of 2.49 on these tasks. See Table 4 in the appendix.

### 4.4 Domain: Causal Reasoning

LaMDA's poorest performance lies in the Causal Reasoning domain, with its domain LRS averaging a 1.34.

In Experiment 10: Causal Gear Task Schulz et al. [2007], in which participants are given evidence and asked to identify the causal relationship between a series of gears, LaMDA's performance varied greatly by condition. In the simplest condition (a machine works when gear A pushes gear B), its LRS was 3.42. When the given causal relationship becomes more complex and unusual (a machine works conjunctively when gears A and B push against each other), LaMDA's accuracy drops to an LRS of 2.02. This suggests that LaMDA is using text information that specifies the most common types of causal relationships. In contrast, children in this study simply relied on the pattern of data they observed and were equally willing to infer unusual or common causal relationships.

LaMDA also faced significant difficulties when faced with causal reasoning tasks using a Blicket Detector mechanism. In Experiment 11: the Blicket Induction Task Gopnik and Sobel [2000], which examines participants' ability to make causal inferences about a novel mechanism using category-based and linguistic labels, LaMDA scored an LRS of 0.98 and 1.06 in each condition.

Across four conditions of Experiment 12: the Disjunctive/Conjunctive Blicket Task Lucas [2014], which assesses LaMDA's proficiency in determining causal relationships using a variety of given evidence, its LRS ranged from a 0.44 at the lowest and 1.12 at the highest. Across tasks, LaMDA is largely unable to track the causal evidence given and make meaningful inferences about causal structures. In its attempt to search its learning data for appropriate responses, LaMDA often responded with outlandish internet links. When asked to identify which objects were Blickets (changed to "Zerpdas" following LaMDA citing existing Blicket papers in its responses), LaMDA confidently, yet incorrectly, identified the Zerpda as a bright blue wig. These failures may indicate that these types of causal learning require more real-world and self-initiated exploration and cannot simply be learned from language. Discovering novel causal structures requires seeking real-life evidence and constantly updating hypotheses about causal structure based on that evidence, rather than simply assuming the causal structures that are most frequently reported in language. See Table 6 in appendix.

## 5 Discussion:

In this work we proposed a developmental lens and framework as a systematic way to evaluate LLMs. We created a battery of classical developmental tasks and converted them to text format for use in LaMDA. We created a rating system called Large Language Models Response Score (LRS) and were able to assign values to quantify LaMDA's ability to respond like a child across crucial developmental domains.

We found that LaMDA responded like a child and provided appropriate answers in experiments involving Social and Moral Understanding (LRS:4.33), perhaps providing evidence that the core of these domains is discovered through language. In contrast, LaMDA's responses were at chance in the Perception (LRS: 2.91) and Theory of Mind (LRS: 2.86) domains, and diverged especially strongly from those of children in Causal Reasoning tasks (LRS: 1.34). These results may show that domains involving capacities majorly outside social reasoning require more real-world and self-initiated exploration and cannot simply be learned from language input, while social and moral reasoning may be rooted more deeply in language.

We find that LaMDA does not seem to follow a human developmental trajectory. It performs more poorly on some tasks that infants solve at a very young age than on others that are solved by humans much later on. Instead, LaMDA's performance reflects differences in how much tasks rely on prior knowledge that may be encoded and transmitted through language, as in the preference for helpers, and how much they rely on the ability to draw inferences from novel patterns of evidence, as in causal learning tasks. However, whatever their origins are, these conceptual milestones will be crucial for genuinely functioning AI systems. Systems will have to understand and apply knowledge of objects, actions and goals, and minds and causality, in much the way that human children do.

We hope that our proposed core developmental domains and the associated battery of developmental tasks can be used by fellow researchers to study other AI models, assess crucial understanding of basic common sense concepts, and gauge their ability to reason using seminal developmental experiments. We also propose that our novel LLM Response Score metric can be used to evaluate other language models, such as GPT, and can be adapted to apply to other key experiments from the psychological literature. In addition the results from this work can influence work in ethical AI by comparing the capacities of agents to social capacities, such as a preference for helpers or an impulse towards altruism, that are arguably foundational for human morality.

All the prompts used in the experiment and output from LaMDA can be found in this document: `https://docs.google.com/document/d/1c-oRRBf-NXRj6rsqNknd3OwMIhst8RPVhBwRgVtvXXo/edit?usp=sharing`

# Appendix

Table 2: Explanation of experiments chosen for this study

| Exp # | Paper, Title, Author, Year | Domain | Summary |
|---|---|---|---|
| 1 | Baillargeon et al. [1985]. | Perception | Probes whether children understand that objects continue to exist while occluded. LaMDA was asked whether a car still existed when it drove behind a curtain. |
| 2 | Wynn [1992] | Perception | Investigates infants' capacity to complete basic addition or subtraction in the real world. After being given evidence of multiple birds sequentially flying behind a curtain, LaMDA was asked to identify the final number of birds. |
| 3 | Gergely [1995] | Perception | Explores when children consider an agent's intentions when interpreting goal-directed behavior. We provided LaMDA evidence of an actor behaving irrationally to reach a goal and probed for whether it identified the behavior as irrational. |
| 4 | Woodward [1999] | Perception | Studies when children attribute behavior to a certain goal. LaMDA was given evidence of an actor consistently reaching for one of two objects. When the positions of the items were switched, LaMDA was asked which item the actor would reach for, the goal object or the decoy. |
| 5 | Wimmer [1983]; Baron-Cohen et al. [1985] | Theory of Mind | Explores when children can attribute false beliefs to actors. In a vignette, LaMDA was introduced to two actors. "Sally" moved "Anne's" toy. LaMDA was privy to the new location while Anne was not. LaMDA was asked both where Anne believes the toy is and its actual location. |
| 6 | Ullman [2023] | Theory of Mind | This task explores LLMs' responses to slight variations of the above task. |
| 7 | Perner et al. [1987] | Theory of Mind | Studies children's ability to attribute false beliefs to actors when the participant is given more information. LaMDA was given evidence that a candy box actually contained pencils rather than candy, then was asked what an ignorant actor would believe was in the candy box. |
| 8 | Hamlin et al. [2013] | Social & Moral Understanding | Explores children's ability to incorporate actor intentionally into social evaluations. LaMDA was given social evidence in which Actor A helps Actor B towards a goal, and Actor C impedes Actor B. Then, LaMDA was asked if it would rather engage with Actor A or C. |
| 9 | Warneken and Tomasello. [2006] | Social & Moral Understanding | Asks whether children consider an agent's goal when determining whether to help the agent. LaMDA was given evidence of an actor coming to an outcome intentionally vs unintentionally. It was then tested on whether it offered help in both conditions or only when the outcome was unintentional. |
| 10 | Schulz et al. [2007] | Causal Reasoning | Investigates children's use of conditional interventions to determine causal structure. LaMDA was introduced to a series of gear mechanisms and asked to intervene in order to determine the causal structure. The two conditions included a simplistic relationship (A turns B) and a complex relationship (A and B turn each other). |
| 11 | Gopnik and Sobel [2000] | Causal Reasoning | Studied children's ability to categorize objects using a novel causal mechanism and labels. LaMDA was given information about a novel machine and was tasked with determining which objects held causal power (i.e. made the machine play music). One condition categorized objects with causal power using novel labels, while the other asked LaMDA to apply the novel labels to the objects with causal power. |
| 12 | Lucas [2014] | Causal Reasoning | Investigated children's abilities to make inferences using evidence of causal relationships. LaMDA was asked to extrapolate about complex causal systems using relational evidence. |

Table 3: LRS Scores for the Perception Domain:

| Exp. | Author | Title | Age | Condition | Score |
|---|---|---|---|---|---|
| 1 | Baillargeon (1985) | Object Permanence | 5 MOS | Original | 3.92 |
| 2 | Wynn (1992) | Number Task | 5 MOS | Original | 1.64 |
| 3 | Gergely (1995) | Intentional Action | 12 MOS | Rational | 4.2 |
| 4 | Woodward (1998) | Violation of Expectation | 8-10 MOS | Standard | 1.98 |

Table 4: LRS Scores for the Theory of Mind Domain:

| Exp. | Author | Title | Age | Condition | Score |
|---|---|---|---|---|---|
| 5 | Wimmer (1983) | Sally Anne Task | 4-5 YR | Original | 3.74 |
| 6 | Ullman (2023) | Theory of Mind | 4-5 YR | 1.Transparent | 2.68 |
| 6 | Ullman (2023) | Theory of Mind | 4-5 YR | 2.Relationship Change | 2.78 |
| 6 | Ullman (2023) | Theory of Mind | 4-5 YR | 3.Trusted Communication | 2.1 |
| 6 | Ullman (2023) | Theory of Mind | 4-5 YR | 4.Additional Actor | 2.4 |
| 7 | Perner (1987) | False Belief | 3 YR | Smarties/Pencils | 3.66 |

Table 5: LRS Scores for the Social and Moral Understanding Domain:

| Exp. | Author | Title | Age | Condition | Score |
|---|---|---|---|---|---|
| 8 | Hamlin (2013) | Helper/Hinderer | 6-10 MOS | Successful Helper/Hinderer | 5.0 |
| 9 | Warneken (2006) | Altruism and Intention | 18 MOS | 1.Constraint: Out of Reach | 4.28 |
| 9 | Warneken (2006) | Altruism and Intention | 18 MOS | 2.Constraint: Physical Obstacle | 3.62 |

Table 6: LRS Scores for the Causal Reasoning Domain:

| Exp. | Author | Title | Age | Condition | Score |
|---|---|---|---|---|---|
| 10 | Schulz (2007) | Schulz Gear | 3-5 YR | 1: A Pushes B | 3.42 |
| 10 | Schulz (2007) | Schulz Gear | 3-5 YR | 2: A and B symbiotic | 2.02 |
| 11 | Gopnik (2000) | Blicket Detector | 3-4 YR | 1: Categorization | 0.98 |
| 11 | Gopnik (2000) | Blicket Detector | 3-4 YR | 2: Induction | 1.06 |
| 12 | Lucas (2014) | Blicket Detector | 4-6 YR | 1: Given hypothesis, Conjunctive | 0.76 |
| 12 | Lucas (2014) | Blicket Detector | 4-6 YR | 2: Given hypothesis, Disjunctive | 0.92 |
| 12 | Lucas (2014) | Blicket Detector | 4-6 YR | 3: Not Given hypothesis/Conjunctive | 1.12 |
| 12 | Lucas (2014) | Blicket Detector | 4-6 YR | 4: Not Given hypothesis/Disjunctive | 0.44 |

Table 7: LRS Average Scores per Domain:

| DEVELOPMENTAL DOMAIN | AVG LRS SCORE |
|---|---|
| PERCEPTION | 2.91 (SD:0.16) |
| THEORY OF MIND | 2.86 (SD:0.25) |
| SOCIAL & MORAL UNDERSTANDING | 4.33 (SD:0.33) |
| CAUSAL REASONING | 1.34 (SD:0.48) |